# Refining Time-Space Traffic Diagrams: A Neighborhood-Adaptive Linear Regression Method

Zhihong Yao, Yi Yu, Yunxia Wu, Hao Li, Yangsheng Jiang, Zhengbing He

*Abstract*—The time-space (TS) traffic diagram serves as a crucial tool for characterizing the dynamic evolution of traffic flow, with its resolution directly influencing the effectiveness of traffic theory research and engineering applications. However, constrained by monitoring precision and sampling frequency, existing TS traffic diagrams commonly suffer from low resolution. To address this issue, this paper proposes a refinement method for TS traffic diagrams based on neighborhood-adaptive linear regression. Introducing the concept of neighborhood embedding into TS diagram refinement, the method leverages local pattern similarity in TS diagrams, adaptively identifies neighborhoods similar to target cells, and fits the low-to-high resolution mapping within these neighborhoods for refinement. It avoids the over-smoothing tendency of the traditional global linear model, allows the capture of unique traffic wave propagation and congestion evolution characteristics, and outperforms the traditional neighborhood embedding method in terms of local information utilization to achieve target cell refinement. Validation on two real datasets across multiple scales and upscaling factors shows that, compared to benchmark methods, the proposed method achieves improvements of 9.16%, 8.16%, 1.86%, 3.89%, and 5.83% in metrics including Mean Absolute Error (MAE), Mean Absolute Percentage Error (MAPE), Congestion Matrix Jaccard Similarity Coefficient (CMJS), Structural Similarity Index Measure (SSIM), and Gradient Magnitude Similarity Deviation (GMSD), respectively. Furthermore, the proposed method exhibits strong generalization and robustness in cross-day and cross-scenario validations. In summary, requiring only a minimal amount of paired high- and low-resolution training data, the proposed method features a concise formulation, providing a foundation for the low-cost, fine-grained refinement of low-sampling-rate traffic data.

## I. INTRODUCTION

THE time-space (TS) traffic diagram discretizes road traffic data into grid cells. It uses color coding to show traffic flow parameters like speed, density, or volume [1], [2]. As a vital tool for traffic flow research and applications, the TS diagram depicts segment-level traffic evolution. TS diagram supports applications like identifying bottlenecks [3], capturing bottleneck propagation [4], analyzing collision impacts [5], predicting secondary collisions [6], estimating emissions [7], predicting travel time [8], and reconstructing trajectories [9]. TS diagram abstracts complex traffic dynamics into a matrix structure and provides a key foundation for research and applications. Its quality directly affects analysis accuracy.

However, due to data collection accuracy and cost limitations, existing traffic time-space data are typically low-frequency. These data are collected using sparse detection methods, such as loop detectors or floating car [10], [11]. It should be noted that although video detectors have largely replaced loop detectors and can provide richer visual information, the cross-sectional traffic flow parameter sequences they generate remain temporally low-frequency [12]. Therefore, whether based on fixed-point detection or moving trajectory sampling, the constructed TS diagrams suffer from severely inadequate resolution. The coarse TS diagrams fail to capture the fine details of short-term traffic fluctuations, which restricts their application in traffic management and decision-making. Therefore, obtaining high-resolution TS diagrams has become a key challenge in enhancing the perception capabilities of intelligent transportation systems. [2] explored the feasibility of constructing TS diagrams using non-rectangular parallelogram (nRP) cells, and demonstrated their superiority in travel time estimation. [13] proposed an area-weighted transformation method to transform conventional TS diagrams into nRP-cell-based TS diagrams. [14] proposed the MDRGCN model to reconstruct missing speed data and simulate congestion propagation under high missing data rates, and [15] proposed a traffic state reconstruction GAN model to reconstruct traffic states for road segments with deficient sensors. In recent years, [16] first introduced the concept of "refinement problem of TS diagrams" from an image processing perspective, marking a significant theoretical contribution in this field. Unlike traditional traffic state estimation, which focuses on macro-level interpolation of temporal or spatial dimensions, TS diagram refinement leverages image high-resolution reconstruction techniques to enhance data density. [16] proposed a simple model based on multiple linear regression,

This work was supported in part by the Humanities and Social Science Foundation of the Ministry of Education in China under Grant 25YJCZH339, the National Natural Science Foundation of China under Grant 72471200, in part by the Fundamental Research Funds for the Central Universities under Grant 2682025GH023, and in part by the Sichuan Science and Technology Program under Grant 2025NSFSC2000. *(Corresponding author: Yunxia Wu, Zhengbing He)*

Z. Yao, Y. Yu, Y. Wu, and Y Jiang are with the National Engineering Laboratory of Integrated Transportation Big Data Application Technology, National United Engineering Laboratory of Integrated and Intelligent Transportation, School of Transportation and Logistics, Southwest Jiaotong University, Sichuan 610031, China (e-mail: zhyao@swjtu.edu.cn; yuyi@my.swjtu.edu.cn; yxwu@my.swjtu.edu.cn; jiangyangsheng@swjtu.edu.cn).
H. Li is with the School of Transportation, Changsha University of Science and Technology, Changsha, China (e-mail: lheastwind@126.com).
Zhengbing He is with Faculty of Science and Engineering, University of Nottingham Ningbo China, Ningbo, China (he.zb@hotmail.com).

assuming that the speed value of high-resolution sub-cells within a low-resolution cell can be estimated as a linear combination of the low-resolution cell own and its surrounding eight cells' speed values. This model was tested on various datasets and demonstrated its effectiveness and transferability in improving TS diagram resolution and reconstructing image details with a small-scale training set, outperforming the classical adaptive smoothing interpolation method [17]. However, since this method relies on the global linear correlation assumption, the refinement results tend to exhibit excessive smoothing of edge information.

In intelligent transportation systems, a key research paradigm involves adapting and migrating computer science technologies to solve practical transportation problems[18], [19], [20]. Following this paradigm, we explore the transfer of image processing techniques to the refinement of TS diagrams. In image processing, the development of high-resolution image reconstruction technology began in the 1960s with the concept of linear and spline interpolation proposed by Harris [21]. Over the following decades, it gradually evolved into several key branches: (1) Interpolation-based methods, such as linear interpolation [22] and nonlinear interpolation [23]. These methods are simple and fast but are limited in reconstructing edges and textures due to their reliance on local smoothing assumptions. (2) Reconstruction-based methods, which include frequency domain methods [24] and spatial domain methods [25], [26]. These methods involve the inversion of degradation models to introduce prior knowledge, but they require significant computational power and multiple iterations. (3) Learning-based methods: These methods have evolved from example-based learning [27], neighborhood embedding [28], and sparse representation [29] to deep learning. Deep learning, initiated by SRCNN [30], has achieved accuracy surpassing traditional methods, and has since developed numerous advanced techniques, such as residual learning [31], GANs [32], and attention mechanisms [33]. The core advantage of deep learning lies in its reliance on large-scale, high-precision training data. However, due to the scarcity of high-resolution TS diagrams, the application of deep learning in the TS diagram refinement is limited.

Among learning-based methods, neighborhood embedding was first introduced by Chang et al. [28]. The core idea is that similar low-resolution image patches correspond to similar high-resolution image patches (local similarity assumption). Therefore, if a set of image patches (a neighborhood) identical to a target low-resolution image patch can be found in the training samples, these can be used to infer the high-resolution representation of the target patch. Surprisingly, the local similarity assumption aligns closely with the repetitive local spatiotemporal patterns commonly found in TS diagrams. TS diagrams contain many similar local patches, and finding a neighborhood for each patch becomes easier when highly similar samples are readily available in the training set.

Inspired by the alignment between the local similarity assumption (of neighborhood embedding) and the inherent repetitive local spatiotemporal patterns of TS diagrams, this paper proposes a refinement method for TS diagrams based on neighborhood-adaptive linear regression. The method fully leverages the local patch similarity and repetitiveness assumptions in TS diagrams. Through adaptive neighborhood search and a local linear regression model, the method achieves a transition from low- to high-resolution TS diagrams. Specifically, for each low-resolution cell to be reconstructed and its surrounding 8 cells forming a 3×3 patch, a search is conducted in the training sample set to find similar neighborhood samples. Subsequently, local multiple linear regression parameters are fitted within the neighborhood to predict the values of the 2×2 high-resolution sub-patch in the target cell. This method not only inherits the linear correlation assumption between TS cells [16], maintaining computational simplicity, but also introduces the local adaptability of neighborhood embedding, avoiding the biases of global models. Therefore, the main contributions of this paper include:

(1) The concept of neighborhood embedding is introduced into the refining of TS diagrams. An adaptive neighborhood search mechanism is developed by utilizing the local pattern repetitiveness assumption in traffic flow. This avoids the over-smoothing tendency of the traditional global linear model when dealing with complex traffic dynamics. The method also achieves efficient improvements in refinement accuracy under small training conditions.

(2) A neighborhood-adaptive linear regression model is proposed. Local regression parameters are dynamically fitted within the neighborhood of each low-resolution patch. This allows the capture of their unique traffic wave propagation and congestion evolution characteristics, outperforming the traditional neighborhood embedding method in terms of local information utilization.

(3) A multi-scale and multi-metric systematic validation is carried out. Experiments are conducted on two real trajectory datasets with complementary features: I-24 MOTION and NGSIM. Cross-dataset validation is performed across multiple scales, ranging from macro (kilometer-level, minute-level) to micro (meter-level, second-level). Moreover, the performance is systematically evaluated under different resolutions and upscaling factors. The effectiveness and robustness of congestion recovery, visual similarity, and structural fidelity are comprehensively validated by introducing image structural metrics such as CMJS, SSIM, and GMSD.

(4) The mechanisms that affect refinement accuracy are elucidated through local and global linear correlations in TS diagrams. Examining the bias-variance trade-off mechanism in neighborhood-adaptive regression demonstrates the superiority of local models in handling heterogeneous patterns, thereby strengthening the theoretical foundation for refinement of TS diagrams.

The remainder of this paper is organized as follows. Section 2 presents the methodology, proposing the neighborhood-adaptive linear regression-based refinement method for TS diagrams. Section 3 conducts a case study, testing and comparing the method using datasets of different scales. Section 4 discusses the underlying principles of the method and issues related to parameter settings. Section 5 concludes the paper and outlines future work.

## II. METHODOLOGY

TS diagrams exhibit significant local similarity and repetitiveness, meaning numerous patches with similar traffic patterns exist at different spatiotemporal locations. Therefore, if two low-resolution patches are identical, their corresponding high-resolution patches also share similarities, and the mapping relationship between low- and high-resolution versions tends to be consistent. To further characterize this relationship, this paper introduces the concept of a "neighborhood." For a given low-resolution patch, we define its neighborhood as a collection of patches from the training set. These patches are of identical size and exhibit similar spatiotemporal traffic characteristics with the target low-resolution patch.

Based on this, this paper proposes a neighborhood-adaptive linear regression method (NALR). The core procedure follows: search its neighborhood within the training sample set using a 3×3 patch composed of the target cell and its surrounding eight cells as input. Subsequently, a local linear regression model is fitted within this neighborhood by leveraging the correspondence between the low-resolution patches and their high-resolution counterparts. This model is then applied to the target cell to produce a refinement result that aligns with the local traffic characteristics.

Table I lists the main parameters and variables covered in the methodology section.

### TABLE I
### PARAMETERS AND VARIABLES

| Indices | Definition |
|---------|------------|
| $X^s$ | Low-resolution TS diagram to be refined |
| $x_e^s$ | Low-resolution cell |
| $U^s$ | The number of cells in $X^s$ |
| $Y^s$ | Ground-truth high-resolution TS diagram of $X^s$ |
| $X^t$ | Real low-resolution TS diagram |
| $Y^t$ | High-resolution TS diagram of $X^t$ |
| $p_r^t$ | 3×3 low-resolution patch cropping from $X^t$ |
| $PT$ | Low-resolution training sample set composed of low-resolution patches |
| $U^{pt}$ | The total number of cropped patches in $PT$ |
| $p_r^{th}$ | 2×2 high-resolution value of $p_r^t$'s center cell |
| $PTH$ | Complete training sample set |
| $p_e^s$ | 3×3 patch formed by $x_e^s$ and its surrounding cells |
| $p_{e,i}^s$ | Internal cell of $p_e^s$ |
| $p_{r,i}^t$ | Internal cell of $p_r^t$ |
| $k$ | Neighborhood size |
| $N_e$ | Neighborhood set of $p_e^s$ |
| $p_{r,m}^{th}$ | Internal cell of $p_r^{th}$ |
| $w_{m,0}$ | Constant term in the linear relationship expression within the neighborhood |
| $w_{m,i}$ | Regression coefficient in the linear relationship expression within the neighborhood |
| $W_e$ | Linear regression parameter set used to refine $x_e^s$ |
| $y_e^s$ | 2×2 high-resolution sub-patch of $x_e^s$ |
| $y_{e,m}^s$ | Internal cell of $y_e^s$ |
| $\hat{y}_{e,m}^s$ | Predicted value of $y_{e,m}^s$ |
| $\hat{Y}^s$ | Predicted value of $Y^s$ |
| $Y_i^s$ | Internal cell of $Y^s$ |
| $\hat{Y}_i^s$ | Internal cell of $\hat{Y}^s$ |

### A. Data Preparation

Test Data: The low-resolution TS diagram to be refined is denoted as $X^s$, with a resolution of $a\,s \times b\,m$ (where each cell has a spatiotemporal size of $a\,s \times b\,m$). $X^s$ consists of multiple low-resolution cells, i.e., $X^s = \{x_e^s | e = 1,2,\ldots,U^s\}$, where $U^s$ represents the number of cells in $X^s$. The corresponding ground-truth high-resolution TS diagram is denoted as $Y^s$. For a single operation of refinement in this paper, the target factor is 4, which implies a 2× upscale in both spatial and temporal dimensions. Thus, the resolution of $Y^s$ is $a/2\,s \times b/2\,m$.

Training Data: Real low-resolution TS diagram $X^t$, with the same resolution as $X^s$. The corresponding real high-resolution TS diagram $Y^t$, which has the same resolution as $Y^s$. The low-resolution TS diagram $X^t$ is cropped into 3×3 patches. To ensure the smoothness and richness of the training samples, the cropping stride is set to 1×1, resulting in overlapping patches. This yields a low-resolution training sample set $PT = \{p_r^t | r = 1,2,\ldots,U^{pt}\}$, where $U^{pt}$ denotes the total number of cropped patches. For each patch $p_r^t$ in $PT$, the corresponding high-resolution values of its center cell are recorded as $p_r^{th}$ (a 2×2 patch).

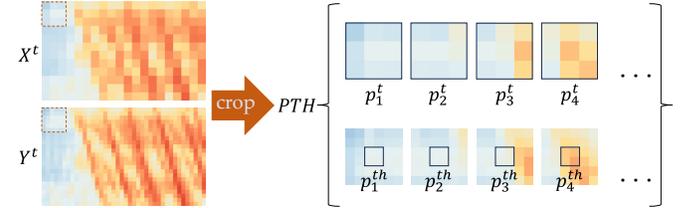

**Fig. 1.** Cropping of low- and high-resolution TS diagrams.

The final training sample set is constructed as $PTH = \{(p_r^t, p_r^{th}) | r = 1,2,\ldots,U^{pt}\}$. Fig. 1 shows an example of obtaining $PTH$ by cropping the low- and high-resolution TS diagrams.

### B. Neighborhood-adaptive Linear Regression

For a given low-resolution cell $x_e^s$ to be reconstructed, the 3×3 patch formed by this cell and its surrounding eight cells are denoted as $p_e^s$. Then, we can calculate the cumulative absolute error (CAE) between $p_e^s$ and each low-resolution patch $p_r^t$ in the training sample set $PTH$ is shown in Eq. (1).

$$CAE(p_e^s, p_r^t) = \sum_{i=1}^{9} |p_{e,i}^s - p_{r,i}^t|. \quad (1)$$

Based on the calculated CAEs, the top $k$ training samples with the smallest distances to $p_e^s$ are selected from $PTH$ to form the neighborhood set $N_e = \{(p_r^t, p_r^{th}) | r = 1,2,\ldots,k\}$. $N_e$ is a subset of $PTH$, and $k$ denotes the neighborhood size, i.e., the number of training samples in the neighborhood. Fig. 2 shows an example of neighborhood search.

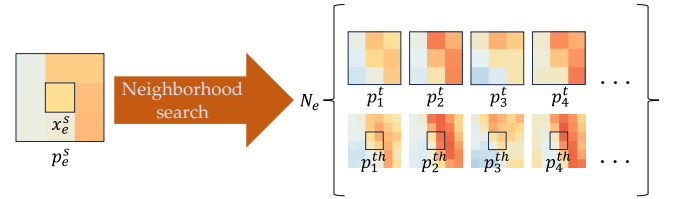

**Fig. 2.** Searching the neighborhood $N_e$ for patch $p_e^s$ where cell $x_e^s$ is located.

Based on the neighborhood search results, the mapping relationship between high- and low-resolutions embedded within the neighborhood is further fitted via neighborhood regression. Specifically, each low-resolution patch $p_r^t$ in the neighborhood set $N_e$ consists of 3×3 cells arranged in the following order: top-left, top-middle, top-right, middle-left, center, middle-right, bottom-left, bottom-middle, bottom-right. The corresponding high-resolution sub-patch $p_r^{th}$ consists of 2×2 cells arranged in the order: top-left, top-right, bottom-left, bottom-right, as illustrated in Fig. 3.

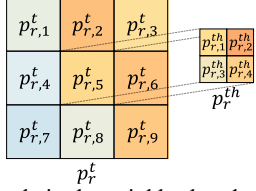

**Fig. 3.** A training sample in the neighborhood.

The objective of neighborhood regression is to fit four sets of multiple linear regression parameters between $p_r^t$ and $p_r^{th}$ (each cell in $p_r^{th}$ individually forms a multivariate linear relationship with the nine low-resolution cells in $p_r^t$). For the m-th dependent variable $p_{r,m}^{th}(m = 1,2,3,4)$ in $p_r^{th}$, the linear relationship can be expressed in Eq. (2).

$$p_{r,m}^{th} = w_{m,0} + \sum_{i=1}^{9} w_{m,i} \cdot p_{r,i}^t, \tag{2}$$

where $p_{r,i}^t$ denotes the value of the cell at the $i$-th position in $p_r^t$, $w_{m,i}$ represents the regression coefficient, and $w_{m,0}$ is the constant term. Using the $k$ samples in the neighborhood, the regression coefficients $W_e = \{w_{m,i}|m = 1,2,3,4; i = 0,1,2,…,9\}$ can be obtained via least squares fitting.

For the low-resolution cell $x_e^s$ to be reconstructed, our objective is to predict the values of the four sub-cells $\{y_{e,m}^s|m = 1,2,3,4\}$ within its corresponding 2×2 high-resolution sub-patch $y_e^s$. The calculation is performed in Eq. (3).

$$\hat{y}_{e,m}^s = w_{m,0} + \sum_{i=1}^{9} w_{m,i} \cdot p_{e,i}^s, \tag{3}$$

where $p_{e,i}^s$ denotes the value of the cell at the $i$-th position in the patch $p_e^s$.

By applying the above steps: neighborhood search, neighborhood regression, and prediction, to each cell in the low-resolution test diagram $X^s$. The entire diagram can be reconstructed. The workflow of the proposed algorithm is summarized in Table II.

TABLE II
THE WORKFLOW OF THE NEIGHBORHOOD-ADAPTIVE LINEAR REGRESSION METHOD

| The neighborhood-adaptive linear regression method |
| --- |
| **For** $x_e^s$ in $X^s$: |
|     get 3×3 patch $p_e^s$ |
|     **For** $p_r^t$ in $PTH$: |
|         calculate $CAE(p_e^s, p_r^t)$ |
|     **End For** |
|     Select the top $k$ samples with the smallest $CAE$ to form $N_e$ of $p_e^s$ |
|     Obtain the multiple linear regression coefficients $W_e$ by fitting from $N_e$ |
|     Predict the 4 high-resolution sub-cells in $x_e^s$ using $W_e$ and $p_e^s$ |
| **End For** |

## III. VALIDATION

### A. Dataset Description

We utilize two representative vehicle trajectory datasets: the Interstate 24 MObility Technology Interstate Observation Network (I-24 MOTION) dataset [34], jointly established by Vanderbilt University's Institute for Software Integrated Systems (ISIS) and the Tennessee Department of Transportation (TDOT), and the Next Generation Simulation (NGSIM) dataset [35]. These two datasets complement each other in terms of temporal and spatial scales, and are used to validate the refinement performance of the TS diagram under different scenarios in this paper. Moreover, both datasets provide real-world vehicle trajectories with rich traffic dynamics and strong temporal and spatial continuity, and have been extensively used in prior studies.

I-24 MOTION is the first large-scale open scientific observation platform for road traffic in the United States, located along an approximately 6.75-kilometer segment of I-24 highway in southeastern Nashville, Tennessee. The I-24 MOTION INCEPTION v1.0 dataset is the first officially released version from this platform. However, due to issues such as camera synchronization and calibration, it contains noise and incompleteness. To address this, Ji et al. [9] developed the VT-tools v1.0, which processes the data by computing the original macroscopic velocity field, generating a smoothed velocity field, and integrating to produce complete trajectories. This results in the I-24 MOTION INCEPTION VT v1.0 dataset, characterized by continuous and smooth velocity and acceleration curves and low noise and error. Therefore, we adopt the I-24 MOTION INCEPTION VT v1.0 dataset to construct the TS diagrams. The dataset includes trajectory data during multiple weekday morning peak periods from November to December 2022. We select multi-lane data from six specific days—November 22, 28, 29, 30, and December 1, 2, 2022—with a spatiotemporal scope of 6000 m×6000 s, as illustrated in Fig. 4.

The I-24 MOTION INCEPTION VT v1.0 dataset encompasses peak-hour scenarios from multiple weekdays and weekends. It comprehensively includes rich traffic features such as congested flow, free flow, and traffic waves, realistically reflecting the typical traffic states on freeways. The I-24 MOTION INCEPTION VT v1.0 dataset is used to construct large-scale TS diagrams at a kilometer-minute resolution to validate the algorithm's refinement performance at a macroscopic scale.

The NGSIM dataset is one of the most widely used public fine-grained traffic trajectory datasets. It includes high-precision microscopic data such as vehicle position, velocity, acceleration, and lane changes. We select two subsets from it: the I-80 freeway and the US-101 highway. Multi-lane spatiotemporal data with an extent of 1600 s×400 m was extracted from the I-80 dataset, and data with an extent of 2400 s×600 m was extracted from the US-101 dataset, as shown in Fig. 5.

Both subsets can depict vehicle operational characteristics at a meter-second resolution, enabling the construction of high-resolution TS diagrams. They primarily validate the algorithm's refinement performance at a fine-grained scale.

### B. Error Evaluation

We employ Mean Absolute Error (MAE) and Mean Absolute Percentage Error (MAPE) to measure the direct numerical differences between the reconstructed result $\hat{Y}^s$ and the true high-resolution TS diagram $Y^s$. The calculation methods are shown in Eqs. (4) and (5).

$$MAE(Y^s, \hat{Y}^s) = \frac{1}{z} \sum_{i=1}^{z} |Y_i^s - \hat{Y}_i^s|, \qquad (4)$$

$$MAPE(Y^s, \hat{Y}^s) = \frac{1}{z} \sum_{i=1}^{z} \left| \frac{Y_i^s - \hat{Y}_i^s}{Y_i^s} \right|, \qquad (5)$$

where $z$ represents the total number of cells in one TS diagram.

Based on the analysis of the image characteristics of TS diagrams, we posit that congested regions (areas where traffic flow speed is below 30 km/h) possess distinct visual features. Furthermore, identifying and localizing congested areas can significantly enhance the practical engineering application value of TS diagrams. Therefore, the accuracy of reconstructing these congested regions is adopted as one of the metrics for evaluating the refinement quality of TS diagrams. Specifically, cells with traffic flow speeds below 30 km/h in the reconstructed TS diagram and the corresponding true high-resolution TS diagram are marked as congested cells. Based on this, binary congestion matrices (where 1 indicates a congested cell and 0 otherwise) are constructed, as shown in Eqs. (6) and (7).

$$J_i = \begin{cases} 1, \text{if } Y_i^s < 30 \text{ km/h} \\ 0, \text{otherwise} \end{cases}. \qquad (6)$$

$$\hat{J}_i = \begin{cases} 1, \text{if } \hat{Y}_i^s < 30 \text{ km/h} \\ 0, \text{otherwise} \end{cases}. \qquad (7)$$

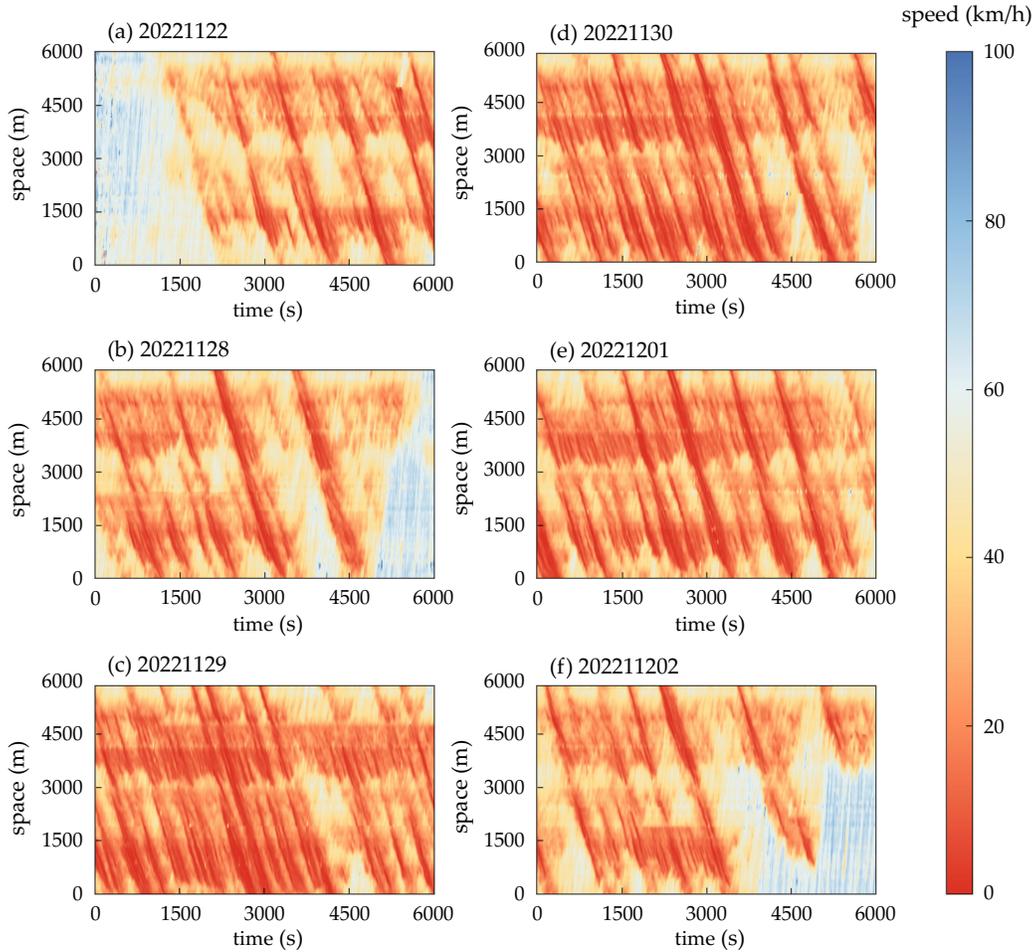

**Fig. 4.** Multi-day scatter trajectory data from the I-24 MOTION INCEPTION VT v1.0 dataset.

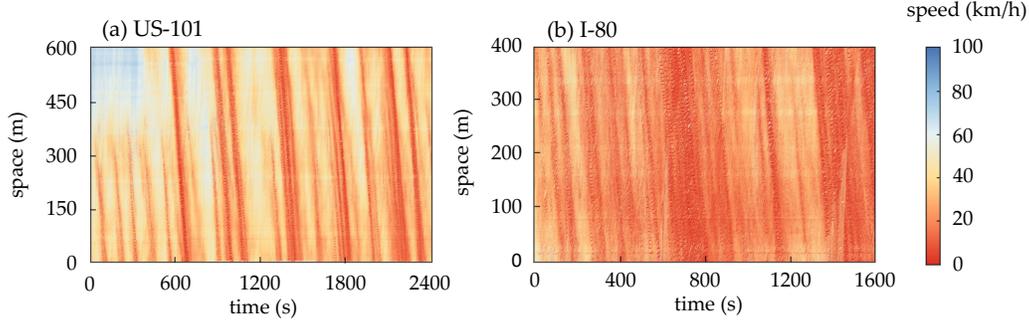

**Fig. 5.** Scatter trajectory data from the US-101 and I-80 datasets in NGSIM.

By comparing the similarity between the binary congestion matrices $J$ and $\hat{J}$, the accuracy of reconstructing congested regions in the refinement task can be quantitatively assessed. We utilize the Jaccard Similarity Coefficient (i.e., Intersection over Union - IoU) to measure the similarity between the two binary matrices. It is defined as the ratio of the intersection to the union of two sets. The Congestion Matrix Jaccard Similarity (CMJS) for $Y_i^s$ and $\hat{Y}_i^s$ is given by Eq. (8).

$$CMJS(Y_i^s, \hat{Y}_i^s) = \frac{J_i \cap \hat{J}_i}{J_i \cup \hat{J}_i}. \tag{8}$$

We also employ the Structural Similarity Index Measure (SSIM) to evaluate the visual quality of the refinements. SSIM assesses the visual similarity between two images by calculating components across three dimensions: luminance, contrast, and structure. The SSIM value ranges from -1 to 1, with values closer to 1 indicating higher visual similarity between the two images. The calculation is shown in Eq. (9).

$$\begin{aligned} SSIM(Y^s, \hat{Y}^s) \\ = \frac{(2\mu_{Y^s}\mu_{\hat{Y}^s} + C_1)(2\sigma_{Y^s\hat{Y}^s} + C_2)}{(\mu_{Y^s}^2 + \mu_{\hat{Y}^s}^2 + C_1)(\sigma_{Y^s}^2 + \sigma_{\hat{Y}^s}^2 + C_2)}, \end{aligned} \tag{9}$$

where $\mu_{Y^s}$ and $\mu_{\hat{Y}^s}$ are the mean values of $Y^s$ and $\hat{Y}^s$, respectively, used to measure luminance. $\sigma_{Y^s}^2$ and $\sigma_{\hat{Y}^s}^2$ are the variances of $Y^s$ and $\hat{Y}^s$, respectively, used to measure contrast. $\sigma_{Y^s\hat{Y}^s}$ is the covariance between $Y^s$ and $\hat{Y}^s$, used to measure structural similarity. $C_1$ and $C_2$ are stabilization constants to prevent division by zero, typically set as $C_1 = (K_1L)^2$ and $C_2 = (K_2L)^2$. $L$ is the dynamic range of the image (i.e., the difference between the maximum and minimum cell values). For TS diagrams, we set the minimum value to 0 and the maximum to 100 (km/h). $K_1$ and $K_2$ are constants, typically set to 0.01 and 0.03, respectively. In practical computation, the image is divided into small local windows, the SSIM is calculated for each window, and then the values are averaged to obtain the overall SSIM index for the entire image. We use a window size of 7×7. If the spatial or temporal dimension of the TS diagram is smaller than 7, the window size is reduced accordingly.

The gradient structures generated by traffic waves represent distinct image features in TS diagrams. To further evaluate the differences in gradients between the reconstructed results and the true TS diagrams, we employ the Gradient Magnitude Similarity Deviation (GMSD) metric. This metric quantifies the consistency of edge and structural features by comparing the gradient magnitude distributions of the two images.

First, we use the Sobel operator to calculate the gradient magnitudes of $Y^s$ and $\hat{Y}^s$, as shown in Eqs. (10) and (11).

$$\nabla Y_i^s = \sqrt{(G_x * Y^s)_i^2 + (G_y * Y^s)_i^2}, \tag{10}$$

$$\nabla \hat{Y}^s = \sqrt{(G_x * \hat{Y}^s)_i^2 + (G_y * \hat{Y}^s)_i^2}. \tag{11}$$

Here, $*$ denotes the convolution operation, and $\nabla Y_i^s$ and $\nabla \hat{Y}^s$ represent the gradient magnitudes of the $i$-th cell in two TS diagrams, respectively. $G_x$ and $G_y$ are the horizontal and vertical convolution kernels of the Sobel operator [36], as defined in Equations (12) and (13).

$$G_x = \begin{bmatrix} -1 & 0 & 1 \\ -2 & 0 & 2 \\ -1 & 0 & 1 \end{bmatrix}, \tag{12}$$

$$G_y = \begin{bmatrix} -1 & -2 & -1 \\ 0 & 0 & 0 \\ 1 & 2 & 1 \end{bmatrix}. \tag{13}$$

At each cell, the similarity of the gradient magnitudes is calculated as shown in Eq. (14).

$$GMS_i(Y^s, \hat{Y}^s) = \frac{2 \times \nabla Y_i^s \times \nabla \hat{Y}^s + C}{(\nabla Y_i^s)^2 + (\nabla \hat{Y}^s)^2 + C}, \tag{14}$$

where $C$ is a constant ensuring numerical stability, set to $1\times10^{-8}$. The value of $GMS_i$ ranges between $[0, 1]$, with higher values indicating greater similarity in gradient magnitudes. The mean of the gradient magnitude similarities across all cells is computed, as shown in Eq. (15).

$$\mu_{GMS} = \frac{1}{z}\sum_{i=1}^{z} GMS_i(Y^s, \hat{Y}^s). \tag{15}$$

Finally, the GMSD value is calculated as shown in Eq. (16).

$$\begin{aligned} GMSD(Y^s, \hat{Y}^s) \\ = \sqrt{\frac{1}{z}\sum_{i=1}^{z}(GMS_i(Y^s, \hat{Y}^s) - \mu_{GMS})^2} \end{aligned} \tag{16}$$

A lower GMSD value indicates higher similarity in the gradients of the two TS diagrams.

### C. Experimental Design

To comprehensively evaluate the performance of the proposed method in the refinement of the TS diagram, we design multiple sets of experiments. These experiments cover different datasets, varying spatiotemporal cell sizes, and 4× and 16× refinement tasks, aiming to validate the method's generalization capability, robustness, and adaptability to diverse traffic scenarios. The specific experimental design is as follows.

Experiment Set 1: The data from November 22, 28, and 29, 2022, in the I-24 MOTION INCEPTION VT v1.0 dataset is used as the training set, while the data from November 30, December 1, and December 2 is used as the test set. Macroscopic-scale TS diagrams are constructed with spatial resolutions of 1000 m, 500 m, and 300 m, and temporal resolutions of 300 s, 180 s, and 60 s.

Experiment Set 2: The US-101 dataset from NGSIM is used as the training set, and the I-80 dataset is used as the test set. Microscopic-scale TS diagrams are constructed with spatial resolutions of 100 m and 40 m, and temporal resolutions of 40 s and 20 s.

After testing, we find that the refinement performance was optimal when the neighborhood size parameter k was set to 100 for Experiment Set 1 and 400 for Experiment Set 2.

### D. Results

Since our method builds upon the assumption of linear relationships between cells from the global linear regression (GLR) method proposed by [16]—which serves as the direct theoretical predecessor of our method—we adopt GLR as a primary benchmark for comparison. Additionally, given that our method draws inspiration from the fundamental principles of the neighbor embedding (NE) method proposed by [28], a classic example-based method in image super-resolution, we also include NE as a secondary benchmark (our proposed method is referred to as neighborhood-adaptive linear regression, NALR). The parameters of the methods all follow the settings specified in [16] and [28].

TABLE III
AVERAGE NUMERICAL METRICS ON THE TEST SET (NOV 30, DEC 1, 2) FOR EXPERIMENT SET 1

| Cell size | Upscaling factor | MAE | | | MAPE | | | $R^2$ | |
|---|---|---|---|---|---|---|---|---|---|
| | | NALR | GLR | NE | NALR | GLR | NE | NALR | GLR |
| 300 s×1000 m | 4× | 3.287 | 3.604 | 3.655 | 0.150 | 0.160 | 0.166 | 0.8319 | 0.9111 |
| | 16× | 4.634 | 4.966 | 4.944 | 0.239 | 0.249 | 0.258 | 0.8713 | 0.9239 |
| 180 s×1000 m | 4× | 3.307 | 3.401 | 3.238 | 0.159 | 0.153 | 0.152 | 0.7989 | 0.8861 |
| | 16× | 4.315 | 4.524 | 4.772 | 0.231 | 0.229 | 0.258 | 0.8348 | 0.9313 |
| 60 s×1000 m | 4× | 2.533 | 2.788 | 2.713 | 0.118 | 0.127 | 0.126 | 0.8366 | 0.8808 |
| | 16× | 3.290 | 3.674 | 3.502 | 0.172 | 0.185 | 0.182 | 0.8142 | 0.9521 |
| 300 s×500 m | 4× | 2.739 | 3.094 | 3.299 | 0.123 | 0.139 | 0.155 | 0.8343 | 0.9138 |
| | 16× | 3.848 | 4.296 | 4.459 | 0.196 | 0.220 | 0.239 | 0.8656 | 0.9236 |
| 180 s×500 m | 4× | 2.125 | 2.333 | 2.350 | 0.099 | 0.106 | 0.111 | 0.8732 | 0.9136 |
| | 16× | 2.910 | 3.256 | 4.152 | 0.151 | 0.167 | 0.230 | 0.8731 | 0.9361 |
| 60 s×500 m | 4× | 1.511 | 1.697 | 1.697 | 0.072 | 0.080 | 0.081 | 0.8571 | 0.9426 |
| | 16× | 2.001 | 2.294 | 2.169 | 0.105 | 0.119 | 0.115 | 0.8613 | 0.9669 |
| 300 s×300 m | 4× | 2.872 | 3.226 | 3.368 | 0.132 | 0.148 | 0.162 | 0.7768 | 0.8909 |
| | 16× | 4.068 | 4.369 | 4.478 | 0.211 | 0.226 | 0.244 | 0.8122 | 0.9246 |
| 180 s×300 m | 4× | 1.916 | 2.241 | 2.206 | 0.089 | 0.103 | 0.107 | 0.8623 | 0.9011 |
| | 16× | 2.638 | 3.045 | 4.245 | 0.138 | 0.158 | 0.240 | 0.8492 | 0.9337 |
| 60 s×300 m | 4× | 1.257 | 1.420 | 1.460 | 0.065 | 0.069 | 0.073 | 0.8532 | 0.9493 |
| | 16× | 1.596 | 1.837 | 1.800 | 0.089 | 0.098 | 0.097 | 0.8655 | 0.9712 |

TABLE IV
AVERAGE MULTIMODAL METRICS ON THE TEST SET (NOV 30, DEC 1, 2) FOR EXPERIMENT SET 1

| Cell size | Upscaling factor | CMJS | | | SSIM | | | GMSD | | |
|---|---|---|---|---|---|---|---|---|---|---|
| | | NALR | GLR | NE | NALR | GLR | NE | NALR | GLR | NE |
| 300 s×1000 m | 4× | 0.845 | 0.824 | 0.812 | 0.889 | 0.862 | 0.848 | 0.142 | 0.140 | 0.141 |
| | 16× | 0.793 | 0.776 | 0.766 | 0.768 | 0.733 | 0.706 | 0.213 | 0.217 | 0.229 |
| 180 s×1000 m | 4× | 0.846 | 0.841 | 0.853 | 0.881 | 0.877 | 0.882 | 0.158 | 0.144 | 0.141 |
| | 16× | 0.812 | 0.793 | 0.787 | 0.750 | 0.747 | 0.689 | 0.225 | 0.229 | 0.237 |
| 60 s×1000 m | 4× | 0.875 | 0.859 | 0.867 | 0.941 | 0.923 | 0.917 | 0.155 | 0.153 | 0.157 |
| | 16× | 0.839 | 0.826 | 0.831 | 0.842 | 0.806 | 0.797 | 0.224 | 0.227 | 0.234 |
| 300 s×500 m | 4× | 0.876 | 0.850 | 0.833 | 0.923 | 0.895 | 0.872 | 0.128 | 0.134 | 0.145 |
| | 16× | 0.822 | 0.807 | 0.788 | 0.806 | 0.764 | 0.719 | 0.213 | 0.228 | 0.238 |
| 180 s×500 m | 4× | 0.893 | 0.878 | 0.881 | 0.956 | 0.945 | 0.938 | 0.110 | 0.116 | 0.119 |
| | 16× | 0.868 | 0.849 | 0.802 | 0.861 | 0.840 | 0.723 | 0.196 | 0.204 | 0.234 |
| 60 s×500 m | 4× | 0.923 | 0.915 | 0.913 | 0.969 | 0.960 | 0.960 | 0.104 | 0.118 | 0.118 |
| | 16× | 0.908 | 0.893 | 0.899 | 0.891 | 0.867 | 0.868 | 0.192 | 0.204 | 0.202 |
| 300 s×300 m | 4× | 0.867 | 0.846 | 0.831 | 0.902 | 0.877 | 0.854 | 0.146 | 0.161 | 0.172 |
| | 16× | 0.818 | 0.807 | 0.796 | 0.761 | 0.729 | 0.693 | 0.232 | 0.242 | 0.253 |
| 180 s×300 m | 4× | 0.907 | 0.896 | 0.894 | 0.958 | 0.944 | 0.939 | 0.116 | 0.126 | 0.134 |
| | 16× | 0.882 | 0.864 | 0.799 | 0.864 | 0.834 | 0.682 | 0.205 | 0.213 | 0.246 |
| 60 s×300 m | 4× | 0.939 | 0.931 | 0.929 | 0.968 | 0.962 | 0.960 | 0.112 | 0.123 | 0.119 |
| | 16× | 0.929 | 0.918 | 0.917 | 0.895 | 0.876 | 0.876 | 0.188 | 0.201 | 0.196 |

TABLE V
NUMERICAL METRICS FOR EXPERIMENT SET 2

| Cell size | Upscaling factor | MAE | | | MAPE | | | R² | |
|---|---|---|---|---|---|---|---|---|---|
| | | NALR | GLR | NE | NALR | GLR | NE | NALR | GLR |
| 40 s×100 m | 4× | 1.555 | 1.691 | 1.721 | 0.087 | 0.098 | 0.091 | 0.9737 | 0.9737 |
| | 16× | 2.231 | 2.318 | 2.260 | 0.136 | 0.142 | 0.125 | 0.9680 | 0.9443 |
| 20 s×100 m | 4× | 1.510 | 1.437 | 1.568 | 0.086 | 0.083 | 0.086 | 0.9712 | 0.9898 |
| | 16× | 1.995 | 1.962 | 2.030 | 0.120 | 0.120 | 0.115 | 0.9412 | 0.9539 |
| 40 s×40 m | 4× | 1.390 | 1.460 | 1.549 | 0.080 | 0.082 | 0.084 | 0.9505 | 0.9590 |
| | 16× | 2.086 | 2.141 | 2.162 | 0.127 | 0.128 | 0.124 | 0.9207 | 0.9358 |
| 20 s×40 m | 4× | 1.041 | 1.092 | 1.220 | 0.060 | 0.064 | 0.071 | 0.9631 | 0.9495 |
| | 16× | 1.599 | 1.657 | 1.727 | 0.095 | 0.100 | 0.102 | 0.8769 | 0.9492 |

TABLE VI
MULTIMODAL METRICS FOR EXPERIMENT SET 2

| Cell size | Upscaling factor | CMJS | | | SSIM | | | GMSD | | |
|---|---|---|---|---|---|---|---|---|---|---|
| | | NALR | GLR | NE | NALR | GLR | NE | NALR | GLR | NE |
| 40 s×100 m | 4× | 0.958 | 0.957 | 0.947 | 0.922 | 0.922 | 0.911 | 0.161 | 0.153 | 0.169 |
| | 16× | 0.938 | 0.920 | 0.921 | 0.744 | 0.745 | 0.708 | 0.215 | 0.230 | 0.244 |
| 20 s×100 m | 4× | 0.949 | 0.949 | 0.951 | 0.896 | 0.913 | 0.892 | 0.184 | 0.155 | 0.188 |
| | 16× | 0.940 | 0.937 | 0.937 | 0.735 | 0.759 | 0.715 | 0.219 | 0.227 | 0.257 |
| 40 s×40 m | 4× | 0.954 | 0.947 | 0.945 | 0.923 | 0.917 | 0.896 | 0.160 | 0.167 | 0.181 |
| | 16× | 0.942 | 0.929 | 0.929 | 0.726 | 0.711 | 0.672 | 0.233 | 0.238 | 0.265 |
| 20 s×40 m | 4× | 0.962 | 0.962 | 0.964 | 0.940 | 0.929 | 0.906 | 0.134 | 0.138 | 0.166 |
| | 16× | 0.947 | 0.944 | 0.945 | 0.784 | 0.757 | 0.719 | 0.208 | 0.222 | 0.262 |

TABLE VII
IMPROVEMENT RATE OF NALR OVER GLR AND NE ACROSS METRICS IN EXPERIMENT SET 1

| Cell size | Upscaling factor | MAE | | MAPE | | CMJS | | SSIM | | GMSD | |
|---|---|---|---|---|---|---|---|---|---|---|---|
| | | GLR | NE | GLR | NE | GLR | NE | GLR | NE | GLR | NE |
| 300 s×1000 m | 4× | 8.80% | 10.06% | 6.19% | 9.92% | 2.58% | 4.17% | 3.10% | 4.79% | -0.97% | -0.15% |
| | 16× | 6.69% | 6.26% | 4.18% | 7.53% | 2.14% | 3.59% | 4.75% | 8.79% | 1.74% | 6.96% |
| 180 s×1000 m | 4× | 2.75% | -2.14% | -3.96% | -4.12% | 0.63% | -0.75% | 0.47% | -0.19% | -9.46% | -11.74% |
| | 16× | 4.63% | 9.57% | -0.66% | 10.48% | 2.38% | 3.20% | 0.51% | 8.90% | 1.41% | 4.88% |
| 60 s×1000 m | 4× | 9.14% | 6.64% | 7.04% | 6.33% | 1.77% | 0.87% | 1.90% | 2.60% | -1.10% | 1.03% |
| | 16× | 10.47% | 6.06% | 6.71% | 5.16% | 1.57% | 0.91% | 4.49% | 5.68% | 1.19% | 4.18% |
| 300 s×500 m | 4× | 11.45% | 16.96% | 11.54% | 20.59% | 3.06% | 5.18% | 3.10% | 5.87% | 4.77% | 11.90% |
| | 16× | 10.43% | 13.70% | 10.61% | 17.77% | 1.97% | 4.37% | 5.47% | 12.06% | 6.38% | 10.56% |
| 180 s×500 m | 4× | 8.89% | 9.56% | 6.41% | 11.10% | 1.67% | 1.34% | 1.16% | 1.89% | 5.43% | 7.51% |
| | 16× | 10.64% | 29.92% | 9.49% | 34.56% | 2.19% | 8.21% | 2.48% | 19.05% | 3.92% | 16.35% |
| 60 s×500 m | 4× | 10.99% | 10.97% | 9.96% | 11.32% | 0.86% | 1.02% | 0.94% | 0.98% | 12.19% | 12.27% |
| | 16× | 12.76% | 7.72% | 11.73% | 8.09% | 1.68% | 1.04% | 2.80% | 2.72% | 5.60% | 4.83% |
| 300 s×300 m | 4× | 10.98% | 14.72% | 11.01% | 18.57% | 2.46% | 4.35% | 2.87% | 5.57% | 9.49% | 15.11% |
| | 16× | 6.90% | 9.15% | 6.62% | 13.40% | 1.32% | 2.78% | 4.34% | 9.82% | 3.96% | 8.36% |
| 180 s×300 m | 4× | 14.51% | 13.14% | 13.65% | 16.40% | 1.29% | 1.53% | 1.43% | 1.99% | 7.94% | 13.61% |
| | 16× | 13.36% | 37.84% | 12.17% | 42.20% | 2.17% | 10.42% | 3.68% | 26.70% | 3.65% | 16.59% |
| 60 s×300 m | 4× | 11.45% | 13.93% | 6.66% | 11.05% | 0.80% | 1.06% | 0.67% | 0.91% | 9.43% | 6.25% |
| | 16× | 13.10% | 11.30% | 8.35% | 7.75% | 1.20% | 1.36% | 2.16% | 2.16% | 6.63% | 4.27% |

TABLE VIII
IMPROVEMENT RATE OF NALR OVER GLR AND NE ACROSS METRICS IN EXPERIMENT SET 2

| Cell size | Upscaling factor | MAE | | MAPE | | CMJS | | SSIM | | GMSD | |
|---|---|---|---|---|---|---|---|---|---|---|---|
| | | GLR | NE | GLR | NE | GLR | NE | GLR | NE | GLR | NE |
| 40 s×100 m | 4× | 8.05% | 9.68% | 10.69% | 3.46% | 0.11% | 1.10% | -0.10% | 1.13% | -5.12% | 4.88% |
| | 16× | 3.75% | 1.27% | 4.04% | -9.04% | 1.97% | 1.91% | -0.12% | 5.13% | 6.59% | 12.09% |
| 20 s×100 m | 4× | -5.06% | 3.69% | -3.10% | 0.24% | 0.01% | -0.26% | -1.84% | 0.39% | -18.82% | 1.80% |
| | 16× | -1.64% | 1.75% | -0.45% | -4.32% | 0.38% | 0.33% | -3.18% | 2.74% | 3.64% | 14.80% |
| 40 s×40 m | 4× | 4.83% | 10.27% | 2.11% | 5.25% | 0.74% | 0.97% | 0.64% | 3.09% | 4.25% | 11.62% |
| | 16× | 2.58% | 3.55% | 0.14% | -3.13% | 1.37% | 1.37% | 2.17% | 8.14% | 1.96% | 12.18% |
| 20 s×40 m | 4× | 4.65% | 14.65% | 6.22% | 15.13% | 0.02% | -0.19% | 1.09% | 3.72% | 2.41% | 19.32% |
| | 16× | 3.51% | 7.46% | 4.43% | 6.73% | 0.39% | 0.29% | 3.59% | 9.07% | 6.15% | 20.53% |

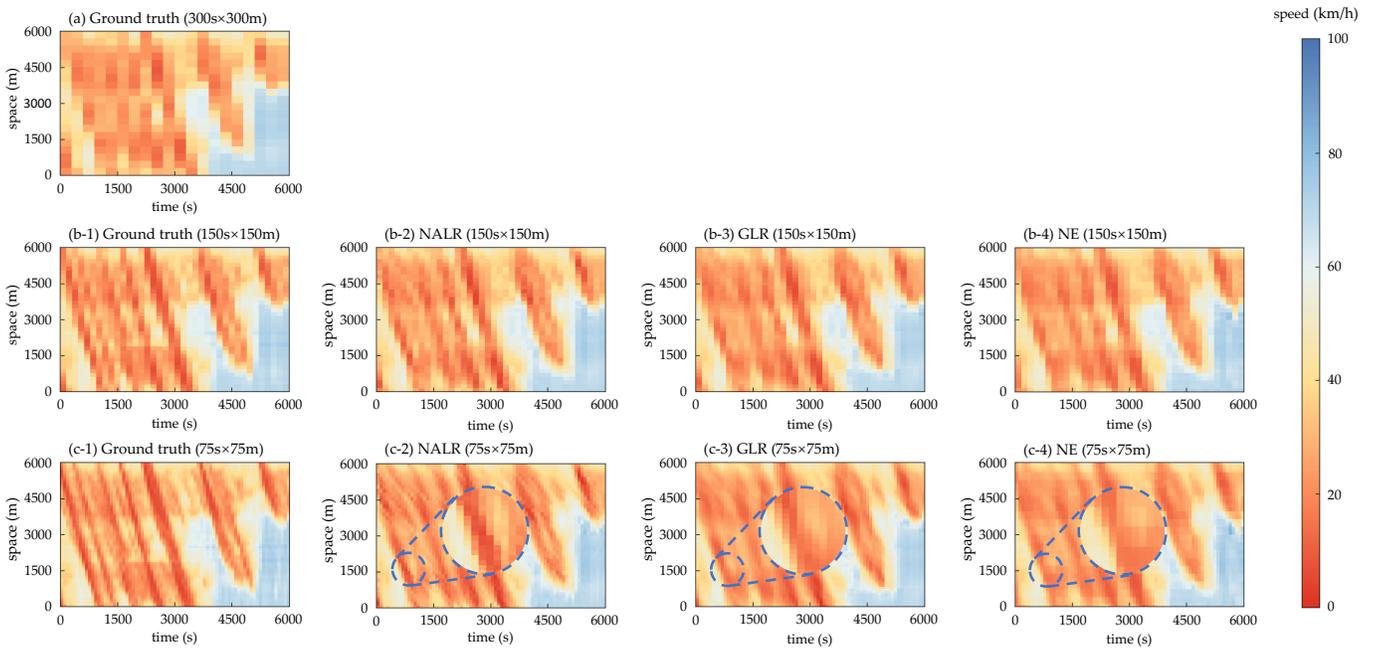

**Fig. 6.** Refinement results for 4× and 16× upscaling of 300s×300m TS diagrams from I-24 MOTION INCEPTION VT v1.0 (December 2, 2022). (a) Low-resolution ground truth to be reconstructed; (b-1) Ground truth corresponding to 4× refinement; (b-2) 4× refinement result using the NALR method; (b-3) 4× refinement result using the GLR method; (b-4) 4× refinement result using the NE method; (c-1) Ground truth corresponding to 16× refinement; (c-2) 16× refinement result using the NALR method; (c-3) 16× refinement result using the GLR method; (c-4) 16× refinement result using the NE method.

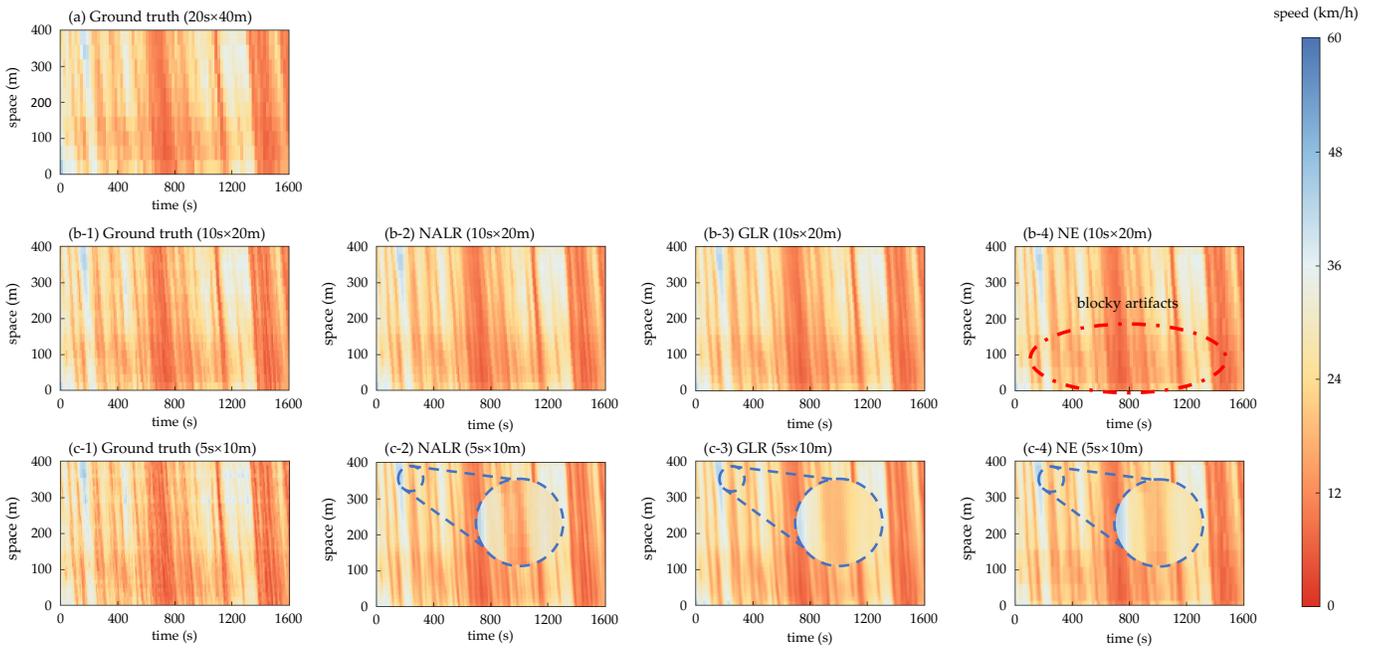

**Fig. 7.** Refinement results for 4× and 16× upscaling of 20s×40m TS diagrams from NGSIM I-80. (a) Low-resolution ground truth to be reconstructed; (b-1) Ground truth corresponding to 4× refinement; (b-2) 4× refinement result using the NALR method; (b-3) 4× refinement result using the GLR method; (b-4) 4× refinement result using the NE method; (c-1) Ground truth corresponding to 16× refinement; (c-2) 16× refinement result using the NALR method; (c-3) 16× refinement result using the GLR method; (c-4) 16× refinement result using the NE method.

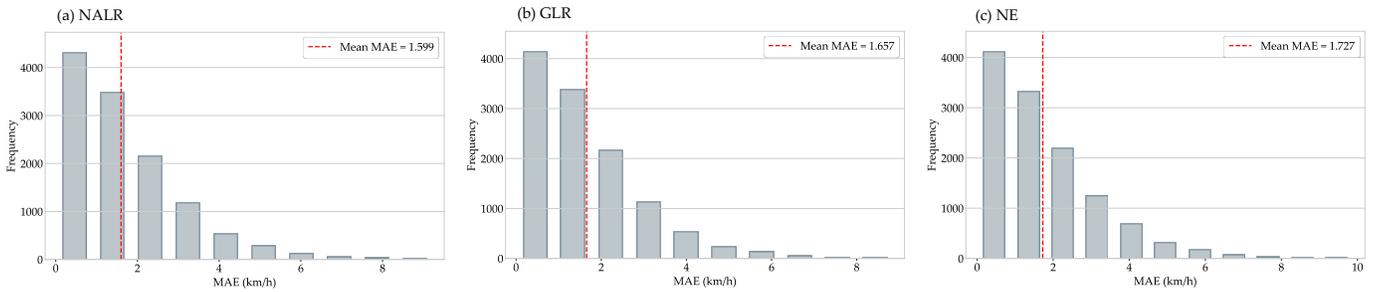

**Fig. 8.** MAE histograms of three methods on 16× upscaling of 20s×40m TS diagrams from NGSIM I-80. (a) NALR; (b) GLR; (c) NE.

The experimental results are grouped by cell size and upscaling factor. Table III and Table IV present the average numerical and multimodal metrics results over the three days in Experiment Set 1. Table V and Table VI present the numerical and multimodal metrics results of Experiment Set 2. Table VII and Table VIII show the improvement rates of NALR over GLR and NE across various metrics in Experiment Set 1 and Experiment Set 2, respectively. An improvement rate indicates that NALR outperforms GLR or NE, while a negative value suggests that NALR underperforms on that specific metric.

Under the macroscopic spatiotemporal scales in Experiment Set 1, NALR achieves approximately 2% to 14% improvement over GLR and 5% to 40% improvement over NE in numerical error metrics such as MAE and MAPE, with only a few exceptions showing negative results. Compared to the GLR method, the improvement rates of CMJS are mostly between 1% and 3%, while compared to the NE method, the improvement can reach over 10%, indicating that the proposed method effectively reduces overall prediction error and maintains the capability to preserve traffic congestion areas.

In terms of SSIM, NALR consistently demonstrates positive improvements across all experimental groups compared with GLR, achieving 1% to 5% enhancement in both 4× and 16× refinement tasks. When compared with NE, the highest improvement rate exceeds 26%. This suggests that NALR outperforms in reconstructing macroscopic structures. However, for GMSD, NALR shows negative results in certain scenarios (e.g., 180 s×1000 m, 60 s×1000 m), yet it still outperforms GLR and NE in most experimental groups.

Overall, the NALR method demonstrates robust performance at macroscopic spatiotemporal scales. Furthermore, we observe that as the spatiotemporal scale becomes finer, the error metrics of NALR consistently decrease. A preliminary analysis suggests that this trend may be attributed to the increased quantity and diversity of samples in the training set as the spatiotemporal cell size reduces. This enhancement facilitates the identification of more similar samples during neighborhood search, a point that will be further validated in subsequent discussions.

The results of Experiment Set 2 also confirm the effectiveness of NALR, demonstrating its superiority across kilometer-minute macroscopic and meter-second microscopic scales. This highlights the method's strong generalization capability and robustness.

Additionally, we note that all coefficients of determination

($R^2$) obtained during the fitting process of the NALR method are lower than those of the GLR method. Nevertheless, NALR still achieves superior refinement performance. The underlying reasons for this phenomenon will be explored in-depth in the following section.

Figs. 6 and 7 demonstrate the refinement effects from representative subsets of Experiment Sets 1 and 2, respectively. The NALR method demonstrates superior visually observable quality compared to GLR and NE, particularly in reconstructing the contours of traffic congestion waves. The NALR method reconstructs sharper and more accurate edges that align more closely with the ground truth, thereby mitigating the over-smoothing issue typically observed in GLR refinement results to a certain extent. Compared with NE, the NALR method achieves superior visual quality by avoiding the blocky artifacts typically induced by per-patch reconstruction.

Fig. 8 presents the MAE histograms of the three methods. It can be observed that, compared with GLR and NE, the MAE distribution of the NALR method is more concentrated around zero with shorter tails (i.e., fewer samples exhibit significant errors).

As for how the GALR method overcomes the limitations of both GLR and NE individually, this will be discussed in subsequent sections.

## IV. DISCUSSION

### A. Why and how NALR is better than GLR and NE

The fundamental advance of the NALR method over GLR lies in its ability to identify a unique set of linear regression coefficients tailored to each cell during refinement, rather than applying a single set of coefficients globally. Although GLR incorporates thresholds to distinguish between free-flow and congested states and performs separate fitting and refinement for each condition, this coarse-grained bifurcation based on global thresholds remains insufficiently specific. Consequently, the GLR method fails to deliver sufficiently precise fitting for cells exhibiting complex transitional states or local heterogeneity, leading to a loss of critical detail in the reconstructed output.

The fundamental superiority of NALR over the traditional NE stems from a transformative innovation in its mechanism for utilizing neighborhoods. The core limitation of NE lies in its reliance on the manifold isometry assumption—that low-resolution and high-resolution patches reside on parallel, structurally similar manifolds. This leads to a process where the weights derived from reconstructing a low-resolution patch within its low-resolution neighborhood are directly transferred to the corresponding high-resolution neighborhood to synthesize the high-resolution output. This operation inherently discards physical consistency, resulting in suboptimal performance when reconstructing traffic time-space diagrams governed by well-defined physical laws.

In contrast, NALR abandons this fragile assumption by replacing indirect transfer with direct mapping. It shifts the core operation from embedding to regression, directly learning the low-to-high resolution mapping from the neighborhood data. Instead of mimicking the way a low-resolution patch is represented by its neighborhood to "hallucinate" the high-resolution patch, NALR learns the underlying functional relationship. Consequently, NALR leverages the concept of neighborhood more fundamentally and effectively, achieving a paradigm shift from reliance on indirect geometric assumptions to direct data-driven learning.

### B. Discussion on Physical Consistency

Although the NALR method is, to some extent, data-driven, its model formulation still maintains a high degree of physical consistency. This study follows the assumption from the GLR model that the relationship between a cell and its surrounding cells is linear—a hypothesis that aligns with the core logic of the Cell Transmission Model (CTM) [37]. CTM is likewise built upon the "discretization of time and space into cells," assuming uniform states within each cell and linear approximate flow-density relationships governing state transitions between adjacent cells. When traffic density is continuous in both time and space, the linear recursive equations of CTM correspond precisely to the discrete representation of linear characteristic propagation in hydrodynamic models, thereby establishing a well-defined linear transfer relationship among local cells. This aligns with the logic of NALR, which fits linear parameters within a local 3×3 spatiotemporal neighborhood. Consequently, the computational process of NALR remains anchored to the physical principles of macroscopic traffic flow, ensuring the physical consistency of the refined results. To experimentally validate this point, we will further analyze how various metrics reflect the consistency with traffic flow physical characteristics.

In terms of congestion formation and queue extension identification, the temporal span of a spatiotemporal congestion region corresponds to the duration of congestion, while its spatial span corresponds to the length of the queue. The CMJS metric quantifies the consistency of spatiotemporal congestion regions, which directly reflects the reconstruction accuracy of congestion formation and queue propagation.

Regarding traffic wave propagation, GMSD calculates the similarity deviation of gradient magnitude in the TS diagram using the Sobel operator, directly reflecting the consistency of traffic waves. The speed of a traffic wave essentially corresponds to the slope of the contour representing abrupt changes in traffic states in the TS diagram, such as the front of congestion or the acceleration wavefront. The gradient serves as a key measure for capturing the intensity and direction of such edge variations. A lower GMSD value indicates a more accurate gradient distribution in the refined results, which in turn implies more precise contours of traffic wave edges and stronger consistency in traffic wave representation.

### C. Discussion on Algorithm Complexity

To comprehensively evaluate the NALR method, we briefly analyze its computational efficiency and complexity. All experiments were conducted on the same computer hardware and operating system environment. The processor was an 11th Gen Intel(R) Core(TM) i5-11320H with a base clock frequency of 3.20 GHz (actual operating frequency 3.19 GHz). The installed RAM capacity was 16.0 GB, with 15.8 GB available. The device was equipped with a 64-bit Windows 11 Home Chinese version (version 24H2, OS build 26100.7171, Windows Feature Experience Pack 1000.26100.265.0).

Table IX presents the average runtimes of the three methods across different datasets and scales. The results show that the GLR method has the shortest runtime in all scenarios due to its simple design of globally uniform regression coefficients, which eliminates the need for per-cell neighborhood search and local fitting. The runtime characteristics of NALR are distinct: in 4× upscaling tasks, its runtime is significantly shorter than that of NE (e.g., 16.708s for NALR vs. 49.258s for NE in Experiment Set 1). However, in 16× upscaling tasks, NALR's runtime exceeds that of NE (e.g., 57.045s for NALR vs. 16.578s for NE in Experiment Set 2). The primary reason is that NALR performs reconstruction in single steps of 4×, whereas NE can perform single-step 16× reconstruction. Therefore, for a 16× reconstruction task on test data of the same resolution, the low-resolution part of the training samples for NALR must match the resolution after a 4× reconstruction, whereas for NE it matches the test data resolution directly. Consequently, although NALR is more efficient with the same number of training samples, in the case of 16× reconstruction in our experiments, NALR got a larger training set than NE, which leads to an increase in computational load.

TABLE IX
AVERAGE RUNTIMES OF THE THREE METHODS ACROSS DIFFERENT
DATASETS AND SCALES

| Experiment set | Upscaling factor | NALR | GLR | NE |
|---|---|---|---|---|
| Experiment set 1 | 4× | 16.708 s | 0.175 s | 49.258 s |
| | 16× | 237.866 s | 2.334 s | 49.606 s |
| Experiment set 2 | 4× | 4.802 s | 0.105 s | 16.579 s |
| | 16× | 57.045 s | 0.219 s | 16.578 s |

Nevertheless, this does not negate the intuitiveness and efficiency of NALR, as it possesses significant efficiency advantages compared to deep learning methods such as SRCNN [30]. SRCNN, as a foundational deep learning super-resolution method, relies on feature extraction and mapping through multi-layer convolutional neural networks. Its computational complexity stems mainly from convolution operations and feature map transformations. Its complexity can be expressed as $O(C_{in} \times C_{out} \times k_{conv}^2 \times H \times W)$, where $C_{in}/C_{out}$ are the input/output channels, $k_{conv}$ is the kernel size, and $H/W$ are the feature map dimensions. Even a simplified SRCNN (e.g., a 3-layer convolutional structure) experiences a significant increase in computational load for convolution operations during $16\times$ upscaling tasks, as feature map dimensions grow quadratically with scale, requiring substantial storage and computation of intermediate feature maps.

The computational core of the NALR method consists of neighborhood search and local linear regression. The neighborhood search phase only requires calculating the CAE between the target $3\times3$ patch and training samples, with complexity $O(k\times9)$ (where the neighborhood size $k$ is set to 100 and 400 in our experiments, and 9 is the feature count of a $3\times3$ patch). The local linear regression phase fits 4 sets of linear parameters (corresponding to the $2\times2$ high-resolution output) via least squares, with complexity $O(k\times9\times4)$. And NALR lacks heavy operations like convolutions or large matrix multiplications, demonstrating notable computational efficiency.

Regarding training, SRCNN requires large-scale paired high- and low-resolution data (typically tens to hundreds of thousands of samples) and optimizes numerous convolution kernel parameters (thousands to tens of thousands for basic SRCNN) via backpropagation over hundreds to thousands of epochs, demanding substantial GPU resources. Furthermore, due to the lack of high-resolution TS diagrams in traffic scenarios, SRCNN is prone to overfitting due to insufficient data.

The training process of NALR does not require complex parameter optimization. It only involves cropping $3\times3$ patches from a small amount of paired data to build the sample set. The training phase involves no backpropagation, gradient descent, or similar operations and can be completed quickly on a CPU without special hardware requirements. Its training complexity is far lower than that of SRCNN.

However, the comparative disadvantages against GLR and NE methods still indicate directions for optimizing the algorithmic efficiency of NALR: improving the efficiency of neighborhood search within training sample sets.

### D. Discussion on Local and Global Linear Assumptions

The NALR method inherits the assumption from the GLR method that relationships between grid cells are linear; however, the resulting $R^2$ values are consistently lower than those of GLR, with some values even below 0.8, while the refinement error decreases. This phenomenon is analyzed as follows.

Beyond local homogeneity, traffic spatiotemporal data often exhibit a degree of heterogeneity: different spatiotemporal regions may adhere to distinct local patterns, making the global linear relationship assumption difficult to fully uphold. The GLR method employs the entire sample set for multiple linear regression, assuming all grid cells share a single set of regression coefficients. This global model achieves high $R^2$ values on the training set because the large-scale samples provide abundant variability, resulting in a substantial total sum of squares (SST); even if the sum of squared errors (SSE) is relatively high, $R^2$ remains elevated. This reflects GLR's strong explanatory power for overall data trends, but it overlooks differences among subgroups, leading to model coefficients that represent a form of "average compromise." Consequently, this introduces systematic bias when predicting specific spatiotemporal grids. For instance, when the target sample lies at the edge of the data distribution or in a minority subgroup (such as the boundary between congested and free flows, corresponding to image edge contours), GLR's global coefficients may amplify extrapolation errors, thereby increasing refinement error.

In contrast, the NALR method adaptively selects similar training samples for each target prediction sample through neighborhood search and regression, thereby constructing local linear models and generating tailored regression coefficients. This approach relaxes the global linear assumption in favor of a local homogeneity assumption, where similar samples follow more consistent linear relationships, even though NALR's $R^2$ values on local subsets are lower (below 0.9 or even 0.8). This primarily occurs because the selected similar samples exhibit lower variability, significantly reducing SST and making $R^2$ more sensitive to SSE—even with good model fits, $R^2$ may decline. However, in terms of predictive performance, NALR mitigates model misspecification risks, avoids biases from global averaging, and ensures that coefficients better align with the local structure of the target sample, thus reducing refinement error.

### E. Analysis of the Impact of Training Set Size

In neighborhood search, the number of samples in the training set influences whether a sufficient number of neighborhood elements similar to the target patch can be found. In the previous experiments, TS diagrams with different resolutions were constructed within certain spatial and temporal ranges. Higher resolutions result in greater grid cells within the spatiotemporal domain, thereby increasing the number of cropped training samples. Fig. 9 illustrates the relationship between the number of training samples and refinement accuracy across experimental groups.

As the number of training samples increases, the refinement errors for both the $4\times$ and $16\times$ refinement groups show a significant decreasing trend, confirming the positive impact of training set size on refinement accuracy.

Fig. 9 also demonstrates that as the number of training samples increases, the decrease in refinement errors gradually slows, a trend consistent with the law of diminishing marginal returns. Once the training sample size becomes sufficient to match target patches with adequately similar neighborhoods, further increasing the number of samples yields progressively weaker improvements in the neighborhood search process.

It is worth noting that beyond the number of training samples, the diversity of the samples also plays a critical role in determining whether sufficiently similar neighborhoods can be matched for target patches, thereby directly influencing refinement accuracy. Consequently, the more comprehensively the training set covers various traffic flow states, the better the refinement performance will be.

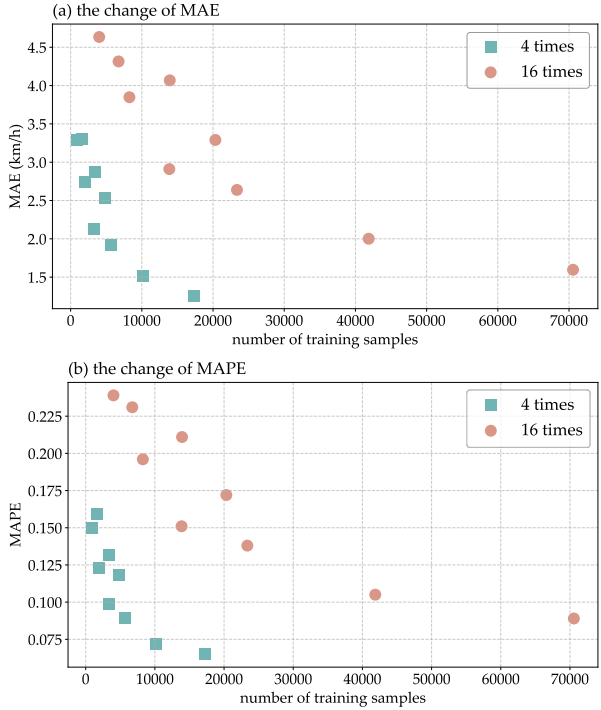

**Fig. 9**. Trends of MAE and MAPE with respect to the number of training samples in Experimental Group 1. (a) Change of MAE with the number of training samples; (b) Change of MAPE with the number of training samples.

### F. Sensitivity Analysis of Neighborhood Size

In the neighborhood regression process, a neighborhood size (k-value) of 100 was used in Experiment Set 1, and a k-value of 400 was adopted in Experiment Set 2. Both values were determined through preliminary traversal attempts. To investigate the impact of this hyperparameter on refinement accuracy and elucidate its underlying mechanism, we provide an in-depth discussion from both theoretical and empirical perspectives.

In the NALR method, k determines the sample size used in local regression. When k is small, the selected neighborhood samples exhibit higher homogeneity, allowing the model to capture better the local linear relationships specific to the target TS cell. This reduces model misspecification risk and systematic bias. However, an insufficient sample size may lead to unstable parameter estimation, increased variance, and heightened sensitivity to noise, thereby amplifying refinement errors. Conversely, a larger k expands the neighborhood to a broader sample space, incorporating more heterogeneous data. This reduces model variance (yielding more robust parameter estimates) but may violate the local homogeneity assumption, resulting in dominance by a global averaging effect and increased bias, consequently raising refinement error. Therefore, a theoretically optimal k value minimizes total prediction error—analogous to k selection in k-NN algorithms.

From an empirical standpoint, refinement experiments were conducted using the December 2nd–60 s×500 m–4× refinement subset from the I-24 MOTION INCEPTION VT v1.0 dataset and the 20 s×40 m–4× refinement subset from the NGSIM dataset. The value of k was traversed from 50 to 1,000 in steps of 50. The variations in error metrics are illustrated in Fig. 10.

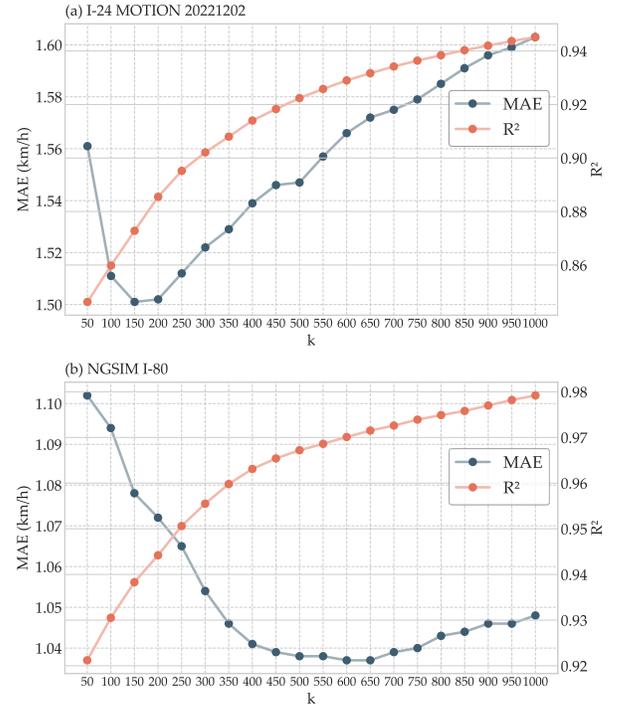

**Fig. 10**. Changes in MAE and R² with respect to k. (a) December 2–60s×500m–4× refinement subset from the I-24 MOTION dataset; (b) I-80–20s×40m–4× refinement subset from the NGSIM dataset.

Empirical results reveal a distinct U-shaped pattern in refinement error. In the I-24 MOTION subset, MSE decreased gradually from 1.561 at k=50 to a minimum of 1.501 at k=150, after which it consistently increased. A similar pattern was observed in the NGSIM subset: MAE decreased from 1.192 at k=50 to a minimum of 1.037 at k=650, followed by a subsequent increase. These trends align with the bias-variance trade-off mechanism discussed earlier, confirming the role of k as a critical hyperparameter in the NALR method.

Furthermore, it was observed that R² gradually increases with larger k, which is consistent with the discussion in Section 4.2 regarding the relationship between R² and local/global linear assumptions.

The optimal k value also varies across datasets due to inherent data characteristics. The I-24 MOTION dataset, with relatively weaker traffic flow heterogeneity (optimal k=150), allows a smaller neighborhood to achieve low variance. In contrast, the optimal k=600 for the NGSIM dataset suggests a more dispersed sample distribution, requiring a larger k to stabilize estimations. This underscores the context-dependent nature of k selection, influenced by the cluster structure of spatiotemporal data and the total sample size.

### G. Sensitivity Analysis of the Size of the Patches Used for Neighborhood Search

Since we are modeling the relationship between the target high-resolution cell and the 3×3 low-resolution cells, the size of the patches used for neighborhood search was set to 3×3 in the neighborhood search step. However, we are naturally curious about how the size of the patches used for neighborhood search may affect the reconstruction results, as it could influence whether more similar neighborhoods can be found for the target patch.

We selected patch sizes of 3×3, 3×1, 1×3, 5×5, 5×1, and 1×5 (in all cases, the target cell is positioned at the exact center of the patch). Experiments were conducted using the December 2nd–60 s×500 m–4× refinement subset from the I-24 MOTION INCEPTION VT v1.0 dataset. The variation in MAE is illustrated in Fig. 11..

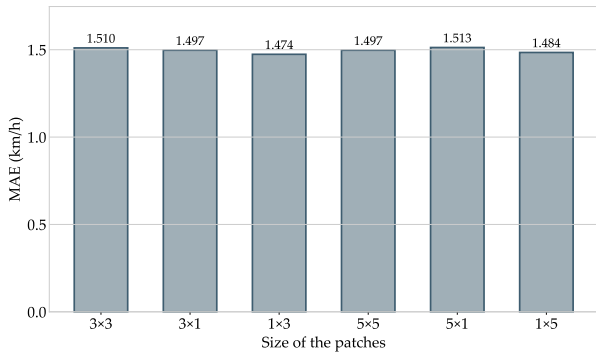

**Fig. 11.** Changes in MAE with respect to the size of the patches (December 2–60s×500m–4× refinement subset from the I-24 MOTION dataset).

It can be observed that, despite testing multiple different patch sizes, MAE did not exhibit significant changes. Therefore, we conclude that the patch size has a negligible impact on the refinement results.

### H. Analysis of the Stability under Perturbations and Data Missing

Due to limitations in detector accuracy and reliability, practical traffic data detection scenarios are often characterized by perturbations and missing data. Therefore, it is necessary to briefly explore the stability under noise perturbations and varying proportions of missing data.

Experiments were conducted using the December 2nd–60 s×500 m–4× refinement subset from the I-24 MOTION INCEPTION VT v1.0 dataset.

Regarding the analysis of noise perturbations, we introduced independent additive Gaussian noise to each cell of the test TS diagram to simulate typical detector measurement errors. This is formulated as: $\tilde{x}_e^s = x_e^s + \epsilon_e, \epsilon_e \sim N(0, \sigma^2)$ , where $\epsilon_e$ represents the Gaussian random noise added to each cell, and $\tilde{x}_e^s$ denotes the corresponding noise-corrupted observation. The standard deviation of the noise was varied within the set $\sigma \in \{0, 0.5, 1, 1.5, 2\}$ $(km/h)$. After noise injection, the TS diagram matrices were truncated to ensure all values remained within the physically valid speed interval—in our experiment, between 0 and 100 km/h. Refinement was then conducted under different noise intensities, and the variations in error metrics are illustrated in Fig. 12.

The results indicate that when $\sigma$ remains within 2 km/h, MAE increases gradually within a tolerable range, implying that low-intensity noise exerts limited influence on the refinement performance. However, as the noise intensity grows further, the reconstruction error exhibits a pronounced rise, suggesting that the stability of the NALR algorithm deteriorates significantly under high-intensity noise interference.

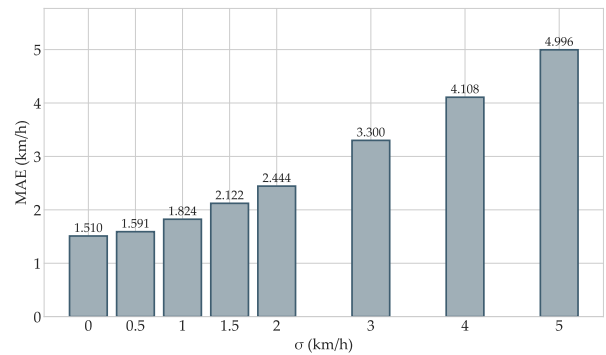

**Fig. 12.** Changes in MAE with respect to $\sigma$ (December 2–60s×500m–4× refinement subset from the I-24 MOTION dataset).

In analyzing data missing, we artificially introduced random missing values at varying proportions into the test TS diagram to simulate the random failure of detectors across both temporal and spatial dimensions. For the missing entries, a smoothing imputation was performed using the mean of non-missing values from the surrounding nine cells. Missing rates were set to 0%, 5%, 10%, 15%, 20%, 30%, 40%, and 50%, respectively. The variation in MAE is illustrated in Fig. 13.

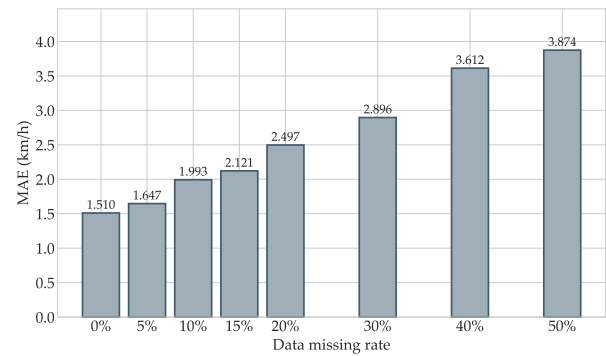

**Fig. 13.** Changes in MAE with respect to data missing rate (December 2–60s×500m–4× refinement subset from the I-24 MOTION dataset).

The results indicate that at a missing rate of up to 20%, MAE shows no significant increase, demonstrating the robustness of the NALR method against data missing.

It should be noted that the adoption of a smoothing imputation approach aligns well with the inherent correlations among adjacent cells, thereby partially mitigating the impact of missing values. Given this observation, we argue that employing such smoothing-oriented data preprocessing strategies can effectively buffer the direct adverse effects of data missingness and noise on the model. This not only supplies the method with more stable and complete inputs but also enhances the robustness in real-world complex data environments.

## V. Conclusion and Future Work

This paper addresses the issue of insufficient resolution in TS diagrams, which leads to missing congestion details and imprecise characterization of propagation features, by proposing a refinement method based on neighborhood-adaptive linear regression.

The NALR method assimilates and adapts the concept of neighborhood embedding, demonstrating that more precise high- and low-resolution mapping relationships exist within local neighborhoods. It overcomes the limitations of existing GLR methods that rely on global linear assumptions, while also outperforming the traditional NE method in terms of local information utilization.

Nevertheless, there remains room for improvement across various stages of the methodology.

(1) Optimization of high-to-low resolution mapping: the linear regression model could be extended to nonlinear approaches, such as lightweight neural network architectures, to balance model complexity and performance effectively.

(2) Optimization of neighborhood search: future work could develop dynamic adjustment strategies based on bias-variance trade-offs to accommodate spatiotemporal characteristics of different datasets and target patches. Furthermore, more advanced feature extraction techniques could be introduced to improve the accuracy of neighborhood search.

(3) Optimization of method applicability: When applied to larger and more complex road networks, the computational efficiency and fitting stability of the proposed method may be compromised due to the expanded neighborhood search scope and increased heterogeneity of local patterns. To address these challenges, hierarchical or partitioned neighborhood search strategies can be developed, or the method can be integrated with graph-based structural modeling, thereby enhancing its general applicability across diverse scenarios.

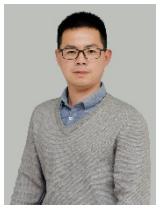

**Zhihong Yao** (Senior Member, IEEE) received the B.S. degree in Transportation Engineering from Southwest Jiaotong University, Chengdu, China, in 2014; and the Ph.D. degree in Transportation Engineering from Southwest Jiaotong University, Chengdu, China, in 2019. He is an associate professor at the School of Transportation and Logistics, Southwest Jiaotong University, Chengdu, China. He also spent one year as a joint doctoral student at the University of Wisconsin-Madison, Madison, WI 53706, USA.

He has authored or co-authored more than 100 academic papers published in the Transportation Research Part C/E, IEEE TRANSACTIONS ON INTELLIGENT TRANSPORTATION SYSTEMS, IEEE TRANSACTIONS ON TRANSPORTATION ELECTRIFICATION, IEEE TRANSACTIONS ON VEHICULAR TECHNOLOGY, IEEE TRANSACTIONS ON INTELLIGENT VEHICLES, etc. His research interests include connected automated vehicles, traffic control, and traffic simulation. He is an Associate Editor for IEEE TRANSACTIONS ON INTELLIGENT TRANSPORTATION SYSTEMS, IEEE TRANSACTIONS ON VEHICULAR TECHNOLOGY, IEEE TRANSACTIONS ON AUTOMATION SCIENCE AND ENGINEERING, IEEE OPEN JOURNAL OF INTELLIGENT TRANSPORTATION SYSTEMS, and an Editorial Board Member of Control Engineering Practice and Sustainable Horizons.

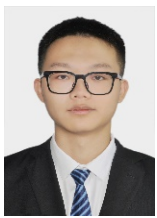

**Yi Yu** received the B.S. degree in Transportation Engineering from Southwest Jiaotong University, Chengdu, China and he is currently pursuing the M.E. degree with the School of Transportation and Logistics, Southwest Jiaotong University, Chengdu, China. His research interests include intelligent transportation systems and connected automated vehicles.

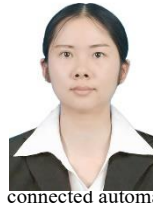

**Yunxia Wu** received the B.S. degree in Environmental Engineering from Anhui Normal University, Chengdu, China, in 2014; and the M.S. degree in Environmental Engineering from Southwest Jiaotong University, Chengdu, China, in 2017. She is a Ph.D. candidate at the School of Transportation and Logistics, Southwest Jiaotong University, Chengdu, China. She has co-authored 20 papers. Her research interests include connected automated vehicles, traffic flow theory, and traffic emissions.

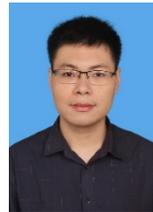

**Hao Li** received the B.Eng. degree from Hebei University of Science and Technology, Shijiazhuang, China, in 2016, and the M.S. degree from Lanzhou Jiaotong University, Lanzhou, China, in 2019. He received the Ph.D. degree from Southwest Jiaotong University, Chengdu, China, in 2023. He is currently a Lecturer with Changsha University of Science & Technology, Changsha, China. He has published more than 10 papers in several reputational academic journals, such as Transportation Letters, Transportation Research Part E, among others. His research interests include autonomous driving and microscopic traffic flow theory.

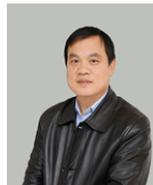

**Yangsheng Jiang** received the B.S. degree in Mechanical Engineering from Yanshan University, Qinhuangdao, China, in 1998; the Ph.D. degree in Transportation Engineering from Southwest Jiaotong University, Chengdu, China, in 2004. He is currently a Professor and Deputy Director of the National Engineering Laboratory of Integrated Transportation Big Data Application Technology, Southwest Jiaotong University, Chengdu, China. His research interests include transportation systems optimization and traffic big data.

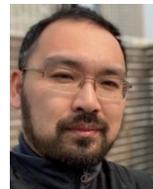

**Zhengbing He** (M'17-SM'20) received the Bachelor of Arts degree in English language and literature from Dalian University of Foreign Languages, China, in 2006, and the Ph.D. degree in systems engineering from Tianjin University, China, in 2011. He was a Post-Doctoral Researcher and an Assistant Professor with Beijing Jiaotong University, China. From 2018 to 2022, he was a Professor with Beijing University of Technology, China. From 2023 to 2025, he was a Research Scientist with Massachusetts Institute of Technology, USA. Presently, he is a Professor with the Faculty of Science and Engineering, University of Nottingham Ningbo China.

His research lies at the intersection of urban mobility, systems engineering, and artificial intelligence, spanning from traditional topics such as traffic flow operations and control, sustainability, and resilience to emerging areas including data-driven modeling, autonomous driving, and large language models. In particular, he is a pioneer and long-term contributor in AI-based transportation modeling and AV-empowered traffic congestion mitigation, and an early innovator in generative AI applications in transportation

He has published more than 180 papers, including over 50 published exclusively in the prestigious journal series of Transportation Research (A/B/C/D/E/F, and 20+ C) and IEEE TRANSACTIONS (10+ TITS), and a Correspondence in Nature, with total citations exceeding 8,000 and H-index of over 40. He was listed as the World's Top 2% Scientists, ranking 67th out of over 30,000 researchers in the field of Logistics and Transportation. He is the Editor-in-Chief of the Journal of Transportation Engineering and Information (Chinese). Meanwhile, he serves as a Senior Editor for IEEE TRANSACTIONS ON INTELLIGENT TRANSPORTATION SYSTEMS, an Associate Editor for IEEE TRANSACTIONS ON INTELLIGENT VEHICLES, a Deputy Editor-in-Chief of IET Intelligent Transport Systems, and an Editorial Advisory Board Member for Transportation Research Part C. His webpage is https://www.GoTraffiGo.com.